\documentclass{article} 
\usepackage{iclr2027_conference,times}
\AtBeginDocument{%
  }

\usepackage{graphicx}
\usepackage{subcaption}
\usepackage{amsmath}
\usepackage{mathtools}
\usepackage{amsthm}
\usepackage{algorithm}
\usepackage{algorithmic}
\usepackage{enumitem}
\usepackage{xcolor}
\usepackage{amssymb}
\usepackage{bm}
\usepackage{wrapfig}

\usepackage{url}
\newlength{\imgH}
\begin{document}

\title{Multi-Task Anti-Causal Learning for Reconstructing Urban Events from Residents' Reports}







\author{Liangkai Zhou\\
Stony Brook University\\
liangkai.zhou@stonybrook.edu \\
\And 
Susu Xu\\
Johns Hopkins University\\
sxu83@jhu.edu \\
\And
Shuqi Zhong \\
Stony Brook University\\
shuqi.zhong@stonybrook.edu\\
\And
Shan Lin\\
Stony Brook University \\
shan.x.lin@stonybrook.edu
}

\iclrfinalcopy

\maketitle

\begin{abstract}
  Many real-world machine learning tasks are anti-causal: they require inferring latent causes from observed effects. 
In practice, we often face multiple related tasks where the structural dependencies are a hybrid of task-invariant and task-specific mechanisms. 
We propose Multi-Task Anti-Causal learning (MTAC), a framework for estimating causes from outcomes and confounders by explicitly exploiting such cross-task invariances. 
MTAC learns a structural equation model (SEM) that factorizes the outcome-generation process into (i) a task-invariant mechanism and (ii) task-specific mechanisms via a shared backbone with task-specific deviations. 
Building on the learned forward model, MTAC performs maximum A posteriori (MAP) based inference to reconstruct causes by jointly optimizing latent mechanism variables and cause magnitudes under the learned structural model.
We evaluate MTAC on the application of urban event reconstruction from resident reports, spanning three tasks: parking violations, abandoned properties, and unsanitary conditions.
On real-world data collected from Manhattan and Newark, MTAC consistently improves reconstruction accuracy over strong baselines, achieving up to 33.04\% MAE reduction and demonstrating the benefits of learning transferable mechanisms across tasks.

\end{abstract}


\section{Introduction}

Many real-world inference problems require estimating latent causes from observed effects, a setting commonly referred to as anti-causal learning. Examples include medical diagnosis~\citep{ahsan2022ml_disease_diagnosis_review}, image classification~\citep{machado2021ml_image_classification_survey}, and urban event reconstruction~\citep{urban_anomaly_analytics}. In these problems, observations are generated through underlying mechanisms that transform causes into observable outcomes. When parts of these mechanisms remain invariant across multiple scenarios or tasks, exploiting such invariance can improve anti-causal inference.


Urban events, such as illegal parking and unsanitary issues, trigger residents' reports through digital platforms (e.g., 311 requests~\citep{nyc311_sr_2020_present}). 
These reports are observations filtered through human decisions, which are shaped by many factors across various types of urban events, including neighborhood socioeconomic status (SES), access to technology, digital literacy etc.~\citep{rotman2014partic, mannarini2010invol, kontokosta2017equity}. 
Consequently, observed reports are not direct measurements of urban events but outcomes filtered through heterogeneous reporting behaviors. Recovering the underlying event distribution from such biased observations therefore requires modeling the report-generation process itself.
We treat the reporting behavior itself as an unobserved mediator between event occurrence and the observed report~\citep{vanderweele2015explanation}.
This mediator is shaped by SES factors of the residents.

Identifying the true causal structure of residents' reporting behavior is inherently difficult, as it involves psychology and social science theory in the complex urban environment. Therefore, we do not aim to construct precise causal models for reporting behavior in this paper. However, the underlying causally-inspired relation structures serve as stronger proxies for reconstructing urban events. Specifically, the same observed report count may arise from fundamentally different combinations of true event incidence and reporting behavior. For example, a neighborhood with frequent events but low reporting activity can generate the same number of reports as a neighborhood with fewer events but high reporting activity. Correlational models~\citep{10027653,Wang2025crime, Butt2025crime, zhou2024crime,Luan2024parking} cannot distinguish between these explanations because they do not explicitly represent the report-generation process. By explicitly modeling how SES and event occurrence jointly generate reports through the reporting behavior, a structural dependency provides an inductive bias that helps disentangle reporting bias from true event incidence.

Moreover, reporting behavior exhibits partial invariance across event types. 
The reporting behavior on different events undergoes a similar psychological path, such as social identity~\citep{mannarini2010invol} and self-efficacy~\citep{rotman2014partic}.
For example, highly educated residents tend to have greater confidence in the effectiveness of formal complaints; consequently, they are more inclined to report problems irrespective of the specific event category~\citep{kontokosta2017equity}.
Meanwhile, the reporting behavior also depends on the event itself.
For example, residents may report events that they believe agencies will act on quickly (e.g., sanitation complaints) at higher rates than events with historically slow responses (e.g., abandoned property intake).
Existing prediction methods~\citep{baseline_urban_report,crime_pred_1,crime_pred_2,parking_pred_1} treat each event type independently and therefore fail to exploit such shared mechanisms.

Exploiting these shared mechanisms requires addressing two challenges. 
First, the causal effects are generated by both task-invariant and task-specific mechanisms. 
In the example of urban event reporting, the reporting behavior of different types of events undergoes a similar psychological path, while also varies depending on the urgency and significance of the event itself.
Therefore, it is critical to disentangle and learn the cross-task invariance.
Second, even if the underlying mechanism is learned, the ultimate objective remains anti-causal: inferring latent event occurrence from observed reports. This requires jointly reasoning about unobserved causes and latent mechanism variables under the learned causal structure. From an anti-causal learning perspective, the partially invariant reporting mechanism provides additional structure that can make this problem more tractable.



To address these challenges, we propose a multi-task anti-causal learning framework, \emph{MTAC}. 
To disentangle task-invariant and task-specific mechanisms, we impose a multi-task structural equation model (SEM) that explicitly represents the mechanism variables as latent factors and decomposes their generative process into task-invariant and task-dependent components, parameterized by separate neural network components. 
To address the second challenge, we develop a MAP-based inference algorithm that estimates the causes by jointly optimizing the causes and the shared mechanism variables under the imposed structural model. 




We evaluate \emph{MTAC} with the urban event reconstruction problem for three types of urban events: abandoned properties, illegally-parked vehicles, and sanitary issues collected through digital platforms (e.g., 311 requests~\citep{nyc311_sr_2020_present}) from New York City and Newark. 

Our contributions are summarized as follows:
\begin{itemize}[leftmargin=*]
    \item We propose MTAC, a structured multi-task SEM whose 
    shared backbone captures the invariant 
    SES to reporting-preference mechanism and whose 
    task-specific heads capture event-specific reporting effects, 
    paired with a MAP-based inversion algorithm that jointly 
    recovers true event counts and latent reporting preferences.
    
    \item 
    We formulate urban event reconstruction as a multi-task anti-causal inference problem in which observed reports arise from true events filtered through partially shared reporting mechanisms influenced by SES and latent reporting preferences.
    
    
    \item We evaluate MTAC on real-world data from Manhattan 
    (parking violations, unsanitary conditions) and Newark 
    (abandoned properties), demonstrating up to 34.61\% MAE 
    reduction over strong anti-causal and multi-task baselines.
\end{itemize}

\section{Related Works}
\label{sec:related}

\subsection{Multi-Task Learning and Multi-Task Causal Discovery}
Multi-task learning (MTL) improves sample efficiency and generalization by sharing representations across related tasks while preserving task-specific components when necessary~\citep{Caruana1997, ple}.
This shared-private design principle is particularly relevant when tasks exhibit both common drivers and task-specific dynamics.
Separately, a growing line of work studies \emph{multi-task} or \emph{multi-environment} causal discovery, aiming to recover causal graphs by pooling data from multiple related distributions or leveraging invariance across tasks/environments~\citep{mt_causal_discovery_jmlr,mt_causal_discovery_aaai,mt_causal_discovery_kdd}.
These methods mainly focus on identifying causal structures (graph recovery) under distribution shifts or heterogeneous datasets.
In contrast, our primary goal is to estimate the cause given the outcome and confounders under a multi-task scenario.

\subsection{Anti-causal Learning and causal prediction}
Predicting causes from observed outcomes is fundamentally different from predicting outcomes from causes.
In anticausal settings, accurate inference typically requires a generative model that captures how causes produce effects, enabling posterior inference over latent causes given observed effects~\citep{anticausal_scholkopf}.
Related problems appear in diagnosis, fault localization, and causal abduction, where one seeks likely explanations of observations under a structural causal model~\citep{CEVAE,cause_prediction_nips,cause_prediction_aaai}.
Nevertheless, much of this literature either (i) does not explicitly leverage the observed outcome information in a structured causal measurement setting, or (ii) relies on strong assumptions such as a observable mediator independent of confounders~\citep{cause_pred_mediator_assumption}.
Our work targets an anti-causal estimation specific to urban event calibration: reports are outcomes generated from events through a latent reporting-preference mechanism.
To address this, we perform maximum a posteriori (MAP) inference of event counts under a learned causal measurement model, effectively ``inverting'' the forward SEM to reconstruct the event distribution from biased reports.

\subsection{Urban Event Prediction from Urban Service Data}
A large body of work studies urban event prediction, including crime incidents~\citep{crime_pred_1, crime_pred_2, crime_pred_3}, parking violations \citep{parking_pred_1}, flooding~\citep{baseline_urban_report}, and other localized urban disorders, from spatiotemporal signals and urban service data such as calls-for-service and 311-style complaints.
Common modeling paradigms include spatiotemporal statistical models, point processes, and, more recently, deep learning approaches such as recurrent models and graph-based spatiotemporal networks that incorporate neighborhood context, mobility, and built-environment features~\citep{crime_pred_1,crime_pred_2,parking_pred_1}.
These methods have demonstrated strong predictive performance when the observed event logs are a reliable proxy of true underlying occurrences.
However, for many civic issues, the observed ``event'' records are themselves partially generated by resident reporting behavior.
Consequently, purely predictive models that treat reported records as ground truth may produce systematically distorted estimates of the event distribution, especially across regions with different socioeconomic status profiles.

\section{Problem Definition}


\begin{wrapfigure}{r}{0.45\linewidth}
    \centering
    \vspace{-0.6in} %
    \includegraphics[width=\linewidth]{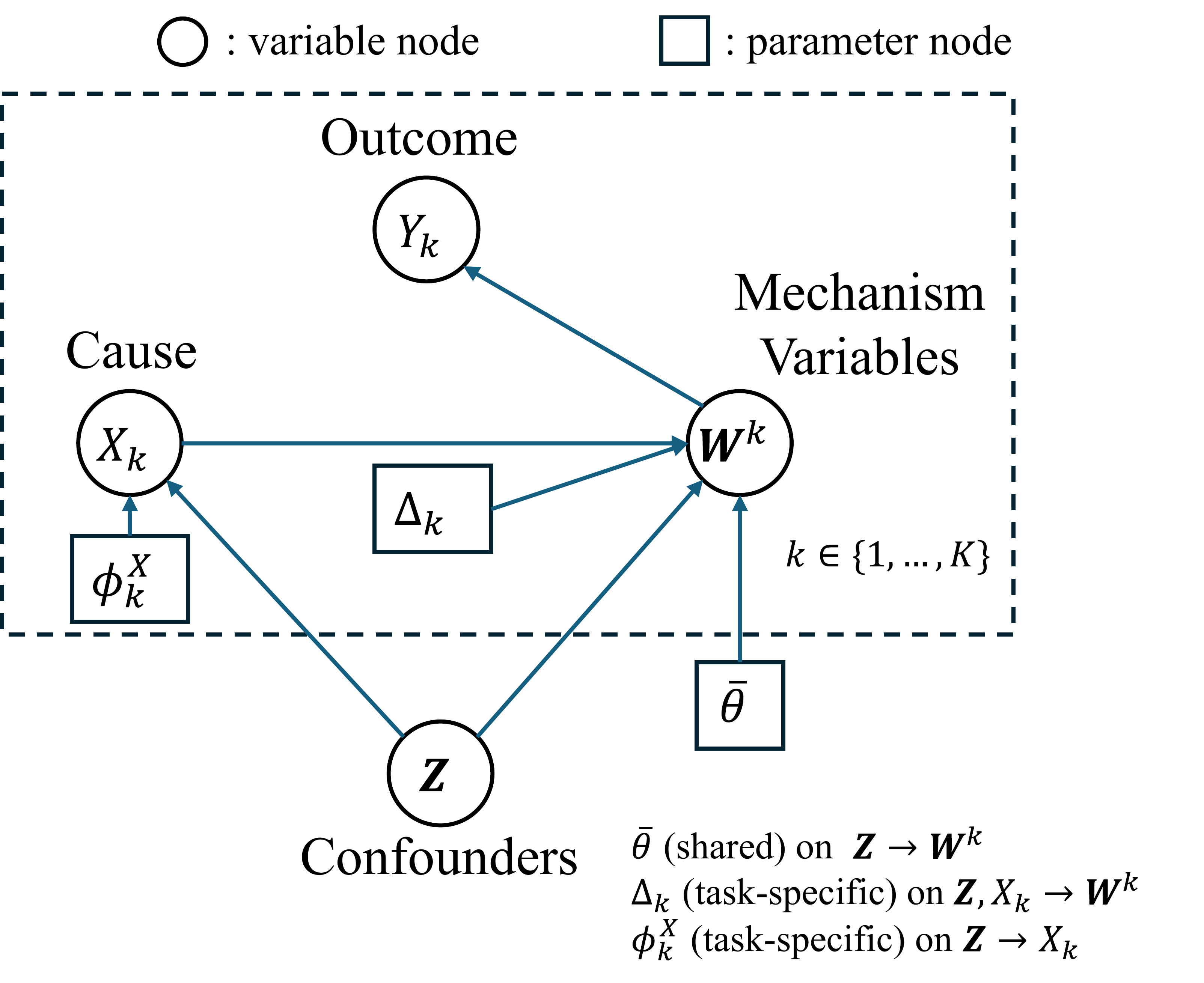}
    \caption{Multi-task causal graph with shared causal mechanism.}
    \label{fig:mt_causal_graph}
\end{wrapfigure}

\label{sec:problem_formulation}

We aim to estimate causes from outcomes and confounders across $K$ related tasks.
In task $k$, let $X_k$ and $Y_k$ denote the cause and the outcome, and let
$\mathbf{Z}=\{Z_l\}_{l=1}^{L}$ denote the observed confounders.
$X_k$ affects $Y_k$ through $M$ latent mechanism variables $\mathbf{W}^k=\{W_i^k\}_{i=1}^{M}$ as mediators,
which are difficult to observe directly.
Following Figure~\ref{fig:mt_causal_graph}, the generative process of task $k$ is
\begin{equation}
    \begin{aligned}
        &X_k \sim p(X_k \mid \mathbf{Z}; \phi^X_k), \\
        &\mathbf{W}^k \sim p(\mathbf{W}^k \mid \mathbf{Z}, X_k; \theta_k), \\
        &Y_k \sim p(Y_k \mid \mathbf{W}^k; \phi^Y_k).
    \end{aligned}
\end{equation}
For each task $k$, we aim to estimate $X_k$ given $Y_k$ and $\mathbf{Z}$,
i.e., $p(X_k \mid Y_k, \mathbf{Z})$.

The process rests on two structural assumptions.
\textbf{(A1) Complete mediation.} 
$X_k$ and $\mathbf{Z}$ affect $Y_k$ only through $\mathbf{W}^k$; there are no direct edges $X_k \to Y_k$ or $\mathbf{Z} \to Y_k$.

\textbf{(A2) Shared mechanism parameters.} 
The mechanism parameters decompose as $\theta_k = \bar{\theta} + \Delta_k$, where $\bar{\theta}$ is shared across tasks and $\Delta_k$ is sparse in the parameters acting on $\mathbf{Z}$.
That is, $\Bar{\theta}$ represents the task-invariant component in the mechanism, while $\Delta_k$ represents the task-dependent component.

Urban event reconstruction is a natural instance of this setting.
A report is filed only when a resident decides to file it, so events and SES
can affect report counts only through this reporting act, which motivates (A1).
The way SES shapes the reporting act reflects properties of residents rather
than of the event type, so we expect the parameters acting on $\mathbf{Z}$ to be
largely shared across event types, while the response to the event itself is
task-dependent, which motivates (A2).
More details are in Appendix~\ref{appendix:case_study}.

\section{Multi-task Structural Model}
\begin{figure}
    \vspace{-0.2in}
    \centering
    \includegraphics[width=0.75\linewidth]{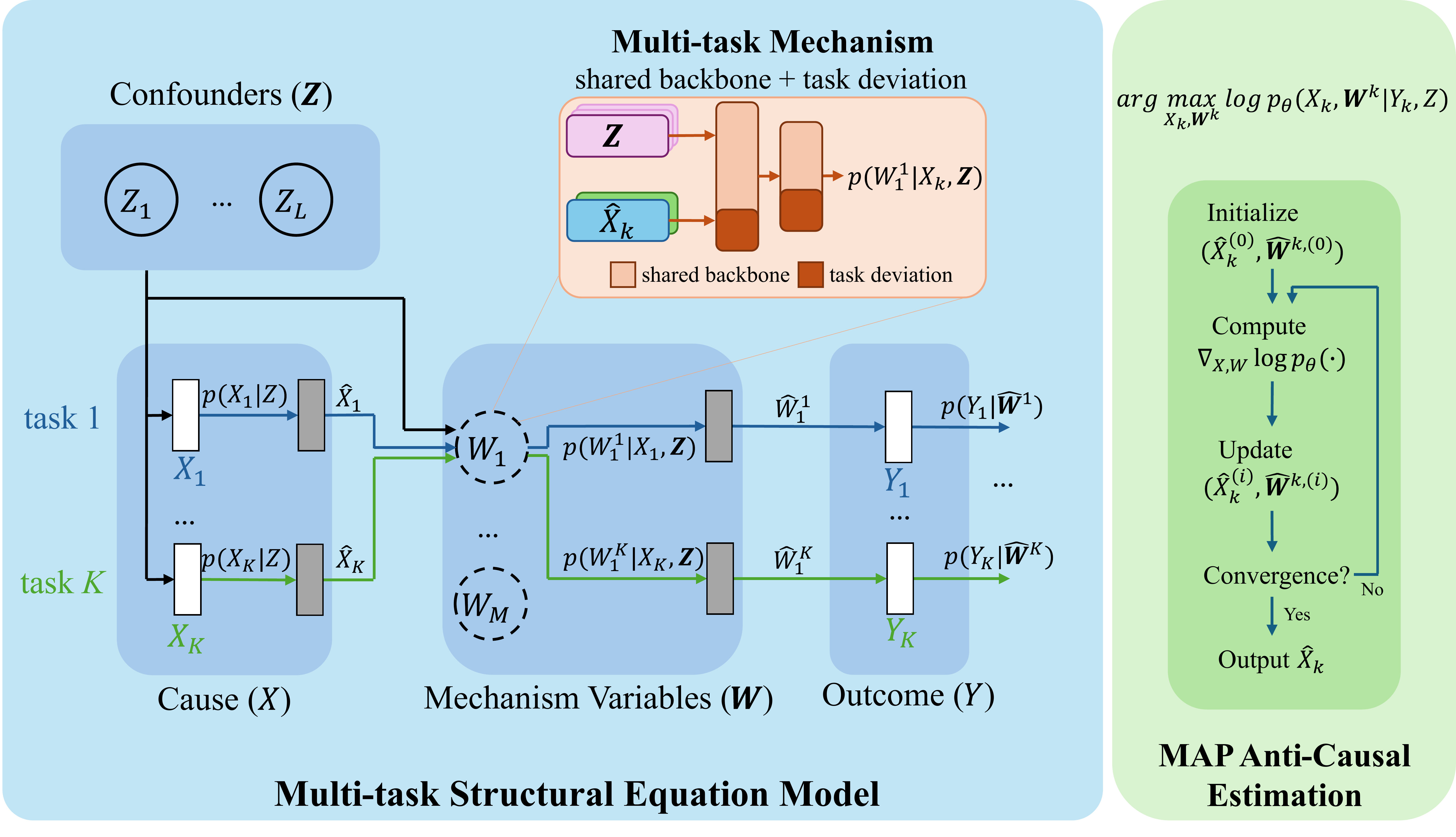}
    \caption{MTAC Framework. The white nodes represent parameterized deterministic neural networks, gray nodes represents drawing samples from the respective distribution. 
    The left panel shows the multi-task structural equation model, in which the mechanism variables W and confounders Z are shared across tasks while causes $X_k$ and outcomes $Y_k$ are task-dependent. 
    The orange box details the multi-task mechanism module for a single mediator $W_1$.
    The right panel illustrates the MAP inference procedure.
    }
    \label{fig:framework}
\end{figure}


To estimate $X_k$ from $Y_k$ for task $k$, it is crucial to model the generation process of $Y_k$, which is a hybrid of task-invariant and task-dependent mechanisms.
Given the heterogeneous datasets for each task, joint training across tasks provides an opportunity to learn a more generalizable and accurate task-invariant structural dependency.
Therefore, we propose MTAC to model the generative mechanism of $Y$ across multiple tasks.
Figure~\ref{fig:framework} illustrates the framework of MTAC.
As shown on the left side of Figure~\ref{fig:framework}, 
a multi-task SEM is constructed to represent the structural dependencies based on Section~\ref{sec:problem_formulation}.
The cause and outcome in each task are modeled as separate variables.
The mechanism variables and confounders are modeled as shared variables across tasks.
The conditional distribution of each variable is modeled with a neural network module, represented as white nodes.
Drawing samples from distributions is represented with gray nodes.
As shown in the orange box in Figure~\ref{fig:framework}, for the mechanism variable $W_1$, we explicitly capture the task-agnostic dependencies from $\mathbf{Z}$ with a task-shared backbone and the task-dependent dependencies from $X_k$ with task deviation module.
To address the anti-causal estimation based on the learned SEM, we employ a MAP-based inference algorithm to estimate the cause $X_k$, as shown on the right side of Figure~\ref{fig:framework}.

\subsection{Multi-task Structural Model}
\label{sec: causal_model}
{

}


We model the structural dependencies with structural equation models. 
Specifically, the distribution of a variable $\mathbf{n}_i$ conditional on its parents $\text{pa}_i$ is modeled as:
\begin{equation}
    \label{eq:sem}
    p(\mathbf{n}_i|\text{pa}_i) \thicksim \mathcal{N}(f^{\mu}_{i}(\text{pa}_i;\theta_i),f^{\sigma}_{i}(\text{pa}_i;\theta_i)).
\end{equation}
$f^{\mu}_{i}(\cdot)$ and $f^{\sigma}_{i}(\cdot)$ fit the mean and variance of $p(\mathbf{n}_i|\text{pa}_i)$ respectively.
$f^{\mu}_{i}(\cdot)$ and $f^{\sigma}_{i}(\cdot)$ are fitted using neural networks with parameter $\theta_i$.
For variables $X$ and $Y$, $f^{\mu}_{i}(\cdot)$ and $f^{\sigma}_{i}(\cdot)$ are modeled with one multi-layer perceptron (MLP) networks with two heads for $\mu$ and $\sigma$ respectively.
Particularly, to explicitly model the task-agnostic and task-dependent dependencies on the mechanism variables $\mathbf{W}$, a multi-task mechanism module is constructed.
Details are discussed in Section~\ref{sec:reporting_model}.

The graph structure is constructed according to Figure~\ref{fig:mt_causal_graph}, where acyclic constraint is naturally satisfied.
Variable $\mathbf{n}_i$'s parents $\text{pa}_i $ is identified with the first layer of $f_{i}(\cdot)$. 
Specifically, let $\bm{\theta}_i^{(1)}$ represent the parameter matrix of the first layer, variable $\mathbf{n}_j$ is $\mathbf{n}_i$'s parent only when the $\bm{\theta}_i^{(1)} $ have sufficiently significant path connecting $\mathbf{n}_j$ to the second layer, i.e., $||\bm{\theta}^{(1)}_{i;[j,:]}||>c$ where $c$ represents a threshold.


\subsection{Multi-task Mechanism Module}
\label{sec:reporting_model}

\begin{wrapfigure}{r}{0.5\linewidth}
    \vspace{-0.1in}
    \centering
    \includegraphics[width=\linewidth]{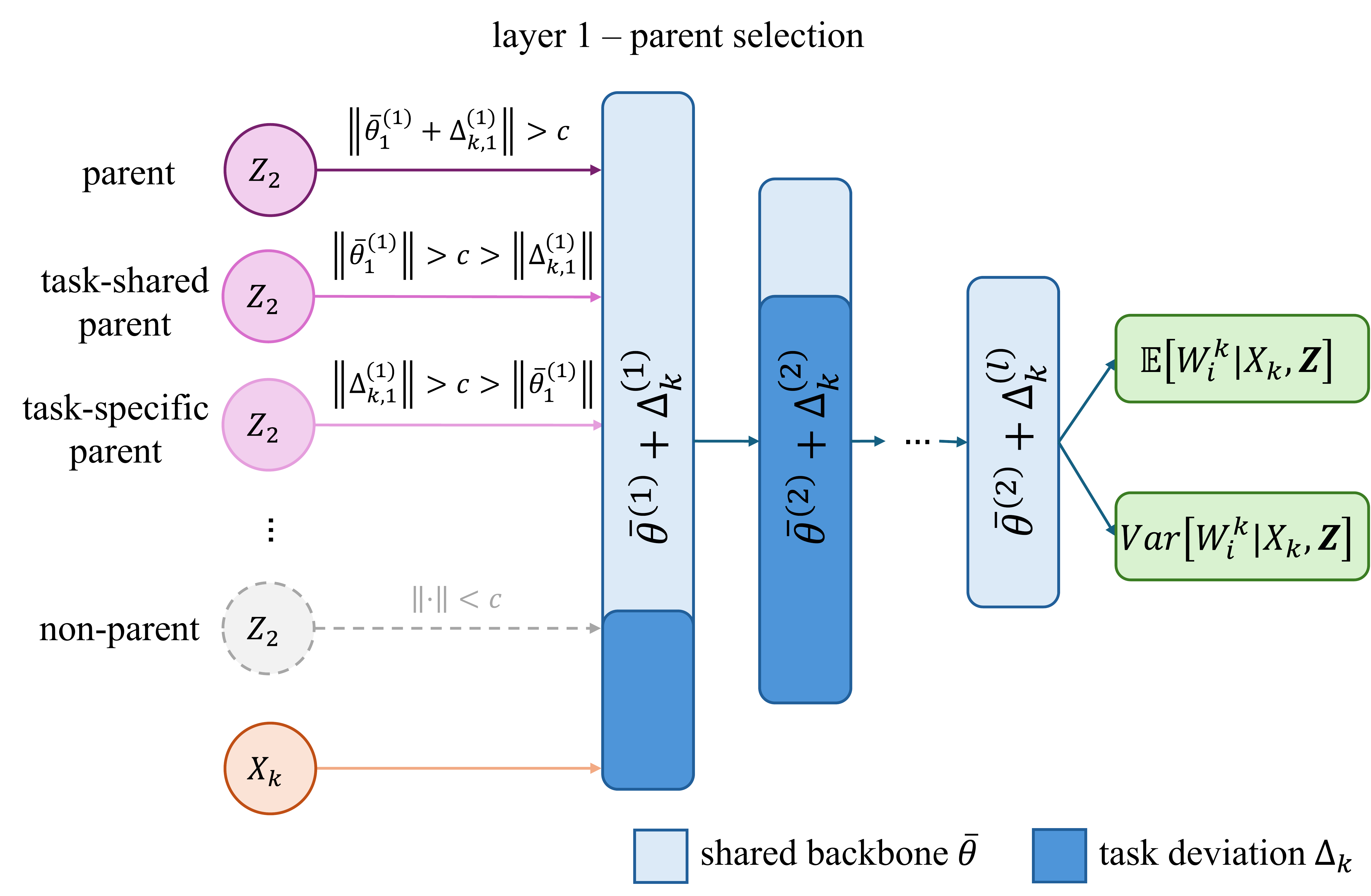}
    \caption{Multi-task SEM for mechanism variable $\mathbf{W}_i$. The norm of first-layer weights decides whether input variables are parents.}
    \label{fig:multi-task_P_SEM}
\end{wrapfigure}

The mechanism variables are generated under both task-dependent effects and task-shared effects.
%
To model both structural dependencies, we construct a multi-task SEM for each mechanism variable, illustrated in Figure~\ref{fig:multi-task_P_SEM}.
For mechanism variable $\mathbf{W}_i$, all confounders $\mathbf{Z}$ and the cause
$X_k$ are supplied to a single network whose parameters at the first $G$ layers are decomposed into a task-shared component and a task-dependent deviation~\citep{regulized_MTL}.
For layer $g \leq G$, we have:
\begin{equation}
    \theta^{(g)}_{i,k} = \bar{\theta}^{(g)}_{i} + \Delta^{(g)}_{i,k},
\end{equation}
As such, that the mechanism variable is generated as
\begin{equation}
    \mathbf{W}^k_i \thicksim
    p(\mathbf{W}^k_i \mid \mathbf{Z}, X_k;\ \bar{\theta}_i, \Delta_{i,k}).
\end{equation}
During training, the larger penalty is applied on the sparsity of $\Delta_{i,k}$ than $\Bar{\theta}_i$.
Details are discussed in Section~\ref{sec:training}.
Since $\Bar{\theta}_i$ is shared by all tasks while $\Delta_{i,k}$ only applies to task $k$, a variable that matters for every task is cheaper to encode once in $\bar{\theta}_{i}$ than separately in each $\Delta^{(1)}_{i,k}$, whereas a variable that matters for a single task is cheaper to encode in that task's deviation alone.
This yields a structural disentanglement of the mechanism into an
invariant and task-dependent component.

\subsection{MAP Anti-Causal Estimation}
\label{sec:map}

The proposed forward structural model enables the estimation of the cause $X_k$ of all tasks from its parental variables directly. 
However, such naive estimation does not leverage the information of the outcome $Y_k$.
Therefore, we develop a MAP-based inference algorithm to estimate the cause by leveraging the outcome information.
For task $k$, the value of $X_k$ can be estimated with its posterior
\begin{equation}
    \label{eq:map_posterior}
    (\hat{X}_k, \hat{\mathbf{W}}^k) = \max_{X_k, \mathbf{W}^k} p_{\theta} (X_k, \mathbf{W}^k | Y_k,  \mathbf{Z}),
\end{equation}
where $\theta$ represents the parameter of the SEM.
Since $p_{\theta} (X_k, \mathbf{W}^k| Y_k, \mathbf{Z}) \varpropto p_{\theta} (X_k, Y_k, \mathbf{W}^k, \mathbf{Z})$, it is equivalent to solve
\begin{equation}
    \label{eq:map}
    (\hat{X_k},\hat{\mathbf{W}}^k) = \arg \max_{X_k,\mathbf{W}^k} \log p_{\theta} (X_k, Y_k, \mathbf{W}^k, \mathbf{Z}).
\end{equation}
Notably, in equation \eqref{eq:map}, we also optimize over the mechanism variables $\mathbf{W}^k$ because they are latent and uncertain. 
Updating them makes the inference more stable.
$p_{\theta}(X_k, Y_k, \mathbf{W}^k, \mathbf{Z})$ can be directly estimated with the forward structural model through the causal order as 
\begin{equation}
    \log p_{\theta}(\mathbf{n}) = \sum_i \log p_{\theta}(\mathbf{n}_i|\text{pa}_i).
\end{equation}
Thus, equation \eqref{eq:map} can be solved with gradient-based optimizers, such as Adam~\citep{kingma2014adam}, since the SEM is constructed using differentiable neural networks.
We first initialize $X_k$ as $p_{\theta} (X_k| \text{pa}( X_k))$ by inferring the structural model.
Similarly, mechanism variables $\mathbf{W}^k$ are initialized by inferring the SEM as well.
Then, $X_k$ and $\mathbf{W}^k$ are optimized by solving equation \eqref{eq:map} using gradient descent methods, while $\theta$ is frozen.
The procedure is illustrated in Algorithm~\ref{alg:map}.

This event inference algorithm bridges the coupled multi-task anti-causal estimation and the shared mechanism.
The mechanism variables $\mathbf{W}^k$ are generated jointly by the $X_k$ and $\mathbf{Z}$.
As a result, the same $Y_k$ can be explained by either (1) a high dependency effect from $X_k$ with a low effect from $\mathbf{Z}$, or (2) a lower effect from $X_k$ with a higher effect of $\mathbf{Z}$.
Taking an example of the urban event reconstruction problem, 
the same amount of reports can indicate a high amount of urban events with low reporting willingness, or a low amount of events with high reporting willingness.
The proposed multi-task structural model supplies a transferable prior on latent mechanism variables, and the MAP inference algorithm leverages it to disentangle reporting bias from true event frequency.
Equation \eqref{eq:map_posterior} can be decomposed to 
\begin{equation}
    \small
    \begin{aligned}
        \arg \max_{X_k, \mathbf{W}^k} 
        &\log p_{\theta} (Y_k|\mathbf{W}_k) + 
        \log p_{\theta} (\mathbf{W}^k | X_k, \mathbf{Z}) \\
        & + \log p_{\theta} (X_k| \mathbf{Z}).
    \end{aligned}
\end{equation}
The first term represents the reporting model, which forces the inferred $\mathbf{W}^k$ to explain the observed outcome $Y_k$.
The second term represents the shared multi-task prior, which regularizes $\mathbf{W}^k$ toward the value learned jointly across multiple tasks.
The third term represents the task-dependent cause prior, which prevents degenerate event estimates that fit $Y_k$ only by inflating $X_k$.
As such, solving equation \eqref{eq:map_posterior} balances the task-dependent and task-shared effects.

\begin{algorithm}[tb]
  \caption{MAP-based event inference}
  \label{alg:map}
  \begin{algorithmic}
    \STATE {\bfseries Input:} Observation of outcome $Y_k$ and confounders $\mathbf{Z}$
    \STATE {\bfseries Output:} Estimation of cause $X_k$
    \STATE Initialize $X_k=\max p_{\theta}(X_k|\mathbf{Z})$.
    \STATE Initialize $\mathbf{W}=\max p_{\theta}(\mathbf{W}|X_k,\mathbf{Z})$.
    \REPEAT
        \STATE Compute joint log probability density $\log p_{\theta}(\mathbf{n}) = \sum_i \log p_{\theta}(\mathbf{n}_i|\text{pa}_i)$ with SEM $\theta$.
        \STATE Update $X_k$ and $\mathbf{W}$ to minimize $-\log p_{\theta }(\mathbf{n})$ with gradient descent method.
    \UNTIL convergence
  \end{algorithmic}
\end{algorithm}

\subsection{Training}
\label{sec:training}


The learning parameters of MTAC are the network parameters $\theta$ of all SEMs.
The graph structure is fixed by the causal order in Figure~\ref{fig:mt_causal_graph}, so no adjacency matrix is optimized and no acyclicity constraint is required.
Instead, the parents of each variable are recovered from the first-layer weights of its SEM, and the loss is designed so that this readout is meaningful.
The loss consists of four terms:
\begin{equation}
    \small
    \mathcal{L} = \mathcal{L}_{\text{nll}}
    + \mathcal{L}_{\text{sp}}
    + \lambda_{\text{div}} \mathcal{L}_{\text{div}}
    + \lambda_{\text{wd}} \mathcal{L}_{\text{wd}}.
\end{equation}

\textbf{Negative log likelihood.}
For variable $\mathbf{n}_i$, the negative log likelihood loss is
\begin{equation}
    \small
    \mathcal{L}^i_{\text{nll}} = - \log p_{\theta}(\hat{\mathbf{n}}_i \mid \text{pa}_i),
\end{equation}
which can be computed directly from the SEM in Eq.~\eqref{eq:sem} without MAP inference.
The confounders $\mathbf{Z}$ are observed exogenous variables and are conditioned on rather than modeled and $\mathbf{W}$ are latent, so no SEM is fitted for them and the overall loss is $\mathcal{L}_{\text{nll}} = \sum_{\mathbf{n}_i \notin \mathbf{Z} \bigcup \mathbf{W}} \mathcal{L}^i_{\text{nll}}$.

\textbf{Sparsity of first-layer weights.}
As discussed in Section~\ref{sec: causal_model}, variable $\mathbf{n}_j$ is identified as a parent of $\mathbf{n}_i$ when the corresponding row of the first-layer parameter matrix has a sufficiently large norm.
To make this readout informative, we penalize the $\ell_2$ norm of each input row, which drives all weights leaving an irrelevant input to zero simultaneously.
For the cause variables, whose SEMs are task-dependent, this gives
\begin{equation}
    \small
    \mathcal{L}^{X}_{\text{sp}} = \lambda_{s}
    \sum_{i \in I_{X}} \sum_{j} \big\| \theta^{(1)}_{i,j,:} \big\|_2 .
\end{equation}
For the mechanism variables, the first-layer weights are decomposed into the task-shared component $\bar{\theta}^{(1)}_{i}$ and the task deviation $\Delta^{(1)}_{i,k}$, and the two are penalized separately:
\begin{equation}
    \small
    \mathcal{L}^{W}_{\text{sp}} = \lambda_{s} \sum_{i \in I_{\mathbf{W}}} \sum_{j \in \mathbf{Z} }
    \big\| \bar{\theta}^{(1)}_{i,j,:} \big\|_2
    + \lambda_{d} \sum_{i \in I_{\mathbf{W}}} \sum_{k} \sum_{j \in \mathbf{Z} }
    \big\| \Delta^{(1)}_{i,k,j,:} \big\|_2 ,
\end{equation}
with $\mathcal{L}_{\text{sp}} = \mathcal{L}^{X}_{\text{sp}} + \mathcal{L}^{W}_{\text{sp}}$.
We set $\lambda_{d} > \lambda_{s}$ so that cross-task sharing is the default and a task deviation is introduced only when the data require it.
The columns corresponding to the cause $X_k$ are excluded from $\mathcal{L}^{W}_{\text{sp}}$, since the dependency of the mechanism on the cause is task-dependent by construction.

\textbf{Non-redundancy of mechanism variables.}
The mechanism variables are latent and share the same parents, so nothing in the likelihood prevents several of them from encoding the same function.
We discourage this with a redundancy penalty on their conditional means.
For a mini-batch of task $k$, let $\tilde{\mu}_i$ denote the batch-standardized value of $f^{\mu}_{i}(\text{pa}_i;\theta)$ for $i \in I_{\mathbf{W}}$, and let $\rho_{ij}$ be the empirical correlation between $\tilde{\mu}_i$ and $\tilde{\mu}_j$.
Following~\citep{bardes2022vicreg}, we penalize the off-diagonal
correlations,
\begin{equation}
    \small
    \mathcal{L}_{\text{div}} = \sum_{k} \sum_{i<j,\ i,j \in I_{\mathbf{W}}} \rho^2_{ij}.
\end{equation}
This term encourages the mechanism variables to span complementary directions rather than collapsing onto a single latent factor.

\textbf{Scale control.}
Under positively homogeneous activations, a network can scale down its first-layer weights and scale up a later layer without changing the function it computes, which would reduce $\mathcal{L}_{\text{sp}}$ at no cost to the fit and render the parent readout vacuous. We therefore apply weight decay to all parameters after the first layer,
\begin{equation}
    \small
    \mathcal{L}_{\text{wd}} = \sum_{i} \sum_{l>1}
    \big\| \theta^{(l)}_{i} \big\|^2_F ,
\end{equation}
which fixes the scale at which the first-layer norms are comparable across variables.


\vspace{-0.05in}
\section{Evaluation}
\vspace{-0.05in}

In this section, we present the empirical results on the urban event reconstruction problem.
These experiments intend to answer the following questions: \textbf{Q1.} Does MTAC improve the performance of urban event reconstruction? (\S\ref{sec:eva:prediction_error})
\textbf{Q2.} Is the task-invariant mechanism valid across tasks? (\S\ref{sec:eva:transfer})
\textbf{Q3.} How much does jointly learning the task-invariant mechanism improve cause estimation? (\S\ref{sec:eva:multi_task})
Appendices~\ref{appendix:ablation} $\sim$ \ref{appendix:sweep} answers several additional question to provide a more comprehensive evaluation on MTAC.

\subsection{Evaluation Setting}
We evaluate our design for the application of urban event reconstruction.
Three types of urban events are considered: 1) parking violations, 2) abandoned properties, and 3) unsanitary conditions. 
The data is collected in New York City and City of Newark. 
Details are discussed in Appendix~\ref{appendix:datasets}.

To our knowledge, no prior method targets the joint setting of multi-task learning, anti-causal estimation, and continuous causes; consequently, we compare against the strongest available baselines from each adjacent line of work and adapt them to our setting where necessary. 

\noindent
\textbf{BSM-UR}~\citep{baseline_urban_report}: A Bayesian spatial latent-variable model that recovers the underlying true event incidence from biased reports.
BSM-UR does not model the causal mechanism of generating resident reports.
It only considers a single type of event.

\noindent \textbf{PLE}~\citep{ple}: A multi-task learning architecture that stacks customized gate-control layers to progressively extract shared and task-dependent features via mixtures of shared/task experts.
PLE does not model the causal relationship between inputs and outputs.

\noindent \textbf{CEVAE}~\citep{CEVAE}: A deep latent-variable causal model that uses a variational autoencoder to account for hidden confounding and enables counterfactual/causal-effect inference (and posterior inference over latent causes) from observational data.
To adapt CEVAE to causes of continuous values, we replace the networks $q(y|t=1,x)$ and $q(y|t=0,x)$ with a network $q(y|t,x)$ that takes the cause $t$ as input.
To estimate the cause, the $p(t|z)$ network in the decoder is inferred.

\noindent \textbf{TEDVAE}~\citep{TEDVAE}: A variational latent-factor causal model that infers hidden factors from observed covariates and disentangles them into instrumental, confounding, and risk components.
To evaluate TEDVAE, we apply the PC algorithm~\citep{PCalgorithm} for causal discovery and categorize the SES variables into these three groups.
To adapt to the causes of continuous values, TEDVAE is modified in the same way as CEVAE.
We infer the auxiliary classifier $q_{w_t}(t|\mathbf{z}_t,\mathbf{z}_c)$ for cause estimation.
TEDVAE does not employ multi-task learning.

To ensure fairness from a data perspective, we train these baselines on the concatenation of all three task datasets, with task indicator as an extra feature.
The event distribution prediction error is evaluated as the difference between the inferred value of the urban event count and the ground-truth.
We measure the prediction error using MAE and MSE.
To validate the contribution of the task-agnostic structural dependencies, we also compare the prediction error when trained separately on each task and when jointly trained across the three tasks.
Moreover, we learn MTAC on two tasks and transfer the task-shared backbone to the other task to validate the task-invariant component of structural dependencies from the SES factors.
Other results are presented in the Appendices~\ref{appendix:ablation} $\sim$ \ref{appendix:sweep}.
Ablation studies are conducted to illustrate the contribution of each component in MTAC (Appendix~\ref{appendix:ablation}).
Policy sensitivity analysis is conducted to understand MTAC's causal utility (Appendix~\ref{appendix:policy_sensitivity}).
A parameter sweep is conducted to understand the multi-path psychological model for reporting behavior (Appendix~\ref{appendix:sweep}).



\subsection{Prediction Error}
\label{sec:eva:prediction_error}
{
\begin{table*}[t]
    \centering
    \small
    \setlength{\tabcolsep}{6pt}
    \begin{tabular}{c|cc|cc|cc}
    \textbf{Model} 
    & \multicolumn{2}{c|}{\textbf{Parking Violation}}
    & \multicolumn{2}{c|}{\textbf{Abandoned Property}}
    & \multicolumn{2}{c}{\textbf{Unsanitary Condition}} \\
    & \textbf{MAE} & \textbf{MSE}
    & \textbf{MAE} & \textbf{MSE}
    & \textbf{MAE} & \textbf{MSE} \\
    \hline
    MTAC     & $\textbf{0.2971}$  & $\textbf{0.1616}$ & $\textbf{0.4163}$ & $\textbf{0.1910}$ & $\textbf{0.3755}$ & $\textbf{0.2501}$ \\
    CEVAE    & $0.3235$ & $0.1925$ & $0.4867$ & $0.2607$ & $0.4091$ & $0.2884$ \\
    TEDVAE   & $0.4315$ & $0.2636$ & $0.4932$ & $0.2583$ & $0.4188$ & $0.2700$ \\
    BSM-UR   & $0.4437$ & $0.2976$ & $0.5438$ & $0.3175$ & $0.3982$ & $0.3026$ \\
    PLE      & $0.3337$ & $0.2174$ & $0.4533$ & $0.2351$ & $0.4101$ & $0.2697$ \\
    \end{tabular}
    \caption{MAE and MSE estimation for MTAC and baselines across three tasks.}
    \label{tab:eval_reconstruction}
    \vspace{-0.2in}
\end{table*}
}

Table~\ref{tab:eval_reconstruction} presents MTAC's prediction error against other baselines.
MTAC outperforms all baselines.
MTAC reduces MAE relative to CEVAE, TEDVAE, and BSM-UR by 8.16\%, 31.15\%, and 33.04\% on parking violations; 14.45\%, 15.59\%, and 23.45\% on abandoned properties; and 8.21\%, 10.34\% and 5.70\% on unsanitary conditions.
The comparison on MSE is also consistent.
This result demonstrates that MTAC learns a more accurate model because the task-agnostic mechanism is learned using data from multiple tasks.
This jointly learned mechanism benefits the reconstruction of all three tasks.
On the other hand, PLE does not model the causal relationship in report generation and only learns the association.
On each dataset, MTAC reduces MAE by 10.97\%, 8.16\%, and 8.44\% compared to PLE, respectively.
MTAC also outperforms PLE in MSE.
This highlights the importance of modeling the structural relationship to avoid spurious associations.

\subsection{Multi-task v.s. Single-task}
\label{sec:eva:multi_task}
To further validate whether jointly learning the shared mechanism contributes to cause estimation, we train and test MTAC on each task separately.
The comparison to jointly-trained MTAC is presented in Table~\ref{tab:eval_MT_ST_comparison}.
When jointly trained on all datasets, the prediction error of MTAC decreases compared to the model trained on each task separately.
MAE and MSE decreased across all tasks: parking violation (0.0722, 0.1331), abandoned property (0.0811, 0.0617), and unsanitary condition (0.0731, 0.0488).
This reduced prediction error in all three tasks demonstrates that jointly learning the task-agnostic structural dependencies leads to a more accurate model.
When learned with single-task datasets, MTAC performed comparably to the baselines on the parking violation and abandoned property tasks and worse than the baselines on the unsanitary condition task.
This is because the baselines are trained using all datasets.
Even with models not designed for multi-task learning, the enlarged dataset already provides a performance gain.

{
}

{
\begin{table}[t]
    \centering
    \small
    \setlength{\tabcolsep}{8pt} 
    \begin{tabular}{l|cc|cc|cc}
    & \multicolumn{2}{c|}{\textbf{Parking Violation}} 
    & \multicolumn{2}{c|}{\textbf{Abandoned Property}} 
    & \multicolumn{2}{c}{\textbf{Unsanitary Condition}} \\
    \textbf{Model} & \textbf{MAE} & \textbf{MSE} & \textbf{MAE} & \textbf{MSE} & \textbf{MAE} & \textbf{MSE} \\
    \hline
    Multi-task  & \textbf{0.2971} & \textbf{0.1616} & \textbf{0.4163} & \textbf{0.1910} & \textbf{0.3755} & \textbf{0.2501} \\
    Single-task & 0.3693 & 0.2947 & 0.4974 & 0.2517 & 0.4486 & 0.2989 \\
    \end{tabular}
    \caption{Performance comparison for MTAC between multi-task training and single-task training.}
    \label{tab:eval_MT_ST_comparison}
    \vspace{-0.15in}
\end{table}

}

\subsection{Task-invariant Mechanism Validation Across Tasks}
\label{sec:eva:transfer}

{
}

\begin{table}[t]
    \centering
    \small
    \setlength{\tabcolsep}{7pt} 
    \begin{tabular}{l|cc|cc|cc}
    & \multicolumn{2}{c|}{\textbf{UC+PV $\to$ AP}} 
    & \multicolumn{2}{c|}{\textbf{UC+AP $\to$ PV}} 
    & \multicolumn{2}{c}{\textbf{PV+AP $\to$ UC}} \\
    \textbf{Model} & \textbf{MAE} & \textbf{MSE} & \textbf{MAE} & \textbf{MSE} & \textbf{MAE} & \textbf{MSE} \\
    \hline
    Full fine-tuned      & \textbf{0.4507} & 0.2313 & 0.3475 & \textbf{0.2502} & \textbf{0.3977} & \textbf{0.2656} \\
    Deviation-only fine-tuned & 0.4631 & \textbf{0.2297} & \textbf{0.3407} & 0.2586 & 0.4106 & 0.2694 \\
    Zero-shot            & 0.7702 & 0.4323 & 0.7699 & 0.5472 & 0.8965 & 0.7518 \\
    Single-task trained  & 0.4974 & 0.2517 & 0.3693 & 0.2947 & 0.4486 & 0.2989 \\
    \end{tabular}
    \caption{Prediction Error of transferred MTAC. The transferred models are trained on two tasks and transferred to the remaining task. PV: parking violation; AP: abandoned property; UC: unsanitary condition.}
    \label{tab:eval_trans}
    \vspace{-0.15in}
\end{table}

To demonstrate that the structural dependencies of the SES on reporting preferences are invariant across urban event types, we also evaluated the reconstruction performance by transferring the model parameters to an unseen task.
We conduct a three-fold evaluation. 
In each fold, we select two tasks and train an MTAC as a teacher model, while a student model is trained on the remaining task.
The shared backbone $\Bar{\theta}$ is transferred from the teacher model to the student model using several transfer learning paradigms:

\noindent \textbf{Zero-shot}: The student model is first trained on the target task.
For all mechanism variables, we replace the student model's shared backbone $\Bar{\theta}$ in the SEMs with that of the teacher model.
Then, the student model is evaluated directly without further fine-tuning.

\noindent \textbf{Deviation-only fine-tuning}: 
Starting from the same teacher initialization, we transfer the parameters of $\Bar{\theta}$ to the student model and freeze this shared module.
We then train only the parameter of target-task deviation module $\Delta$ for mechanism variables and the complete SEMs for other variables) using the target-task training dataset while freezing $\Bar{\theta}$.

\noindent \textbf{Full fine-tuning}: 
For the fully fine-tuned student model, $\Bar{\theta}$ is also optimized after being transferred from the teacher model. 

Table~\ref{tab:eval_trans} presents the reconstruction error of each student model on each dataset.
The performance of a single-task trained version of MTAC is also included as a baseline.
In all three evaluations, both the deviation-only and fully fine-tuned student models outperform the single-task trained model.
Particularly, the deviation-only fine-tuned model uses the exact backbone $\Bar{\theta}$ learned from the other two tasks.
It shows higher accuracy than the single-task trained model in all three folds.
This validates that the structural dependencies from the confounders to the mechanism variables holds invariance across tasks.
When transferring to the abandoned property task, the reduction in both MAE and MSE is most significant compared to the other tasks.
Again, this is due to the limited size of the abandoned property dataset.
Moreover, the zero-shot model performs poorly on all tasks due to heterogeneity across tasks.
The full fine-tuned model and deviation-only fine-tuned model perform similarly in each fold. 
This is because the task-shared backbone captures the task-agnostic patterns and is updated only slightly during full fine-tuning.
The average normalized $L_2$ distance between the task-shared backbone in fully fine-tuned and deviation-only fine-tuned model is 0.064, which represents a small parameter distance.

\vspace{-0.1in}
\section{Conclusion}
\vspace{-0.1in}

This work formulates the multi-task anti-causal learning problem and proposes MTAC to leverage the task-invariant structural dependencies for enhancing cause estimation.
MTAC consists of a multi-task structural equation model while explicitly modeling the task-invariant and task-dependent structural dependencies of latent mechanism variables.
The task-shared and task-dependent mechanisms are explicitly captured using separate neural network modules.
A MAP-based inference algorithm is applied to invert the learned forward SEM for estimating the magnitude of causes from outcomes.
Three urban event reconstruction tasks are evaluated as applications: parking violations, abandoned properties, and unsanitary conditions.
Evaluations show that jointly learning the shared mechanism improves reconstruction accuracy compared with single-task training.
The results also validate the shared structural dependencies across tasks.

\section{AI Use Statement}
In this work, we used generative AI tools to provide feedback on research  methodology or experiments.
We have not used generative AI tools to assist with translation, clean and reformat datasets, support qualitative and thematic data analysis, and interpret results.
Generating synthetic data sets, formulating mathematical claims, providing critical ingredients for proving mathematical claims, and assisting in the writing of proofs are not applicable to this work.
Additionally, we used generative AI tools for brainstorming, creating and editing software code, polishing paper writing to improve readability, and formatting references.
We have reviewed all AI-assisted work.
We have verified the code integrity and language usage of the paper.
We take responsibility for the final content of this work, including text, claims, or artifacts produced with the aid of generative AI.

\section{Reproducibility statement}
We have made several efforts to ensure the reproducibility of our results.
The problem setting and the structural assumptions underlying MTAC are stated in Section~3.
The model architecture, the MAP inference procedure (Algorithm~1), and the complete training objective with all loss terms are described in Sections~4.1--4.4.
Appendix~A.2 lists the data sources for all three tasks, the spatial and temporal aggregation, and the socioeconomic factors used (Table~4).
Appendix~A.3 reports the network architectures, hyperparameters, data splits, normalization, and hardware.
All report and ground-truth data for the Manhattan tasks, as well as the census data, come from publicly available sources cited in Appendix~\ref{appendix:datasets}.
The Newark abandoned-property records were obtained from the City of Newark; we describe their processing in Appendix~\ref{appendix:datasets}.

\bibliography{reference}
\bibliographystyle{iclr2027_conference}

\appendix
\section{Appendix}
\subsection{A Case Study: Urban Event Reconstruction}
\label{appendix:case_study}
\label{sec:urban_event}


Urban event reconstruction is an application where multi-task anti-causal learning being effective.
Residents submit reports upon observing various urban events, such as parking violations, abandoned properties, and unsanitary sites.
This procedure can be understood as a causal generation, where urban events cause residents' reports through a causal mechanism representing their reporting preferences, which are influenced by the residents' socioeconomic status~\citep{kontokosta2017equity}. 
Thus, the causal model for urban report generation is modeled as follows: 1) cause ($X$): number of urban events; 2) outcome ($Y$): number of resident reports; 3) confounders ($Z$): residents' SES; 4) mechanism variables ($W$): residents' reporting preference.

{
}

The reporting act is widely studied in social research~\citep{kontokosta2017equity,Peterson2003psy}. 
The SES factors that shape whether and how a resident files a report can be grouped into five broad categories: 1) financial, 2) educational attainment, 3) race and culture, 4) access to technology, and 5) social environment. 
For example, people with higher educational attainment often trust more that their complaints will be addressed~\citep{Cook2024potholes}. 
Social studies~\citep{mannarini2010invol,rotman2014partic} identify a corresponding set of psychological pathways through which SES influences the reporting behavior, such as social identity, trust in government, self-efficacy, emotion, and digital literacy. 
Motivated by this social-science evidence for a multi-dimensional reporting-behavioral representation, we construct multiple latent mechanism variables. 
Because ground-truth measurements of these latent factors are not available at scale in urban data, MTAC does not enforce or claim a one-to-one correspondence between individual $W_i$ dimensions and named psychological variables. 
Individual dimensions are treated as unlabeled latent factors on the reporting-act pathway, and their contribution to the model is evaluated through end-to-end reconstruction of $X$ rather than through direct semantic validation. 

Notably, these psychological pathways are independent of the urban event type.
Therefore, learning a unified reporting preference model across multiple tasks is more generalizable and may benefit each task.
The causal effect of urban event occurrence is task-dependent, reflecting the heterogeneous reporting propensities of residents toward different categories of urban incidents~\citep{KONTOKOSTA2021102503}.
As such, the multi-task causal mechanism design in MTAC would explicitly capture both the task-shared and task-dependent causal effects of residents' reporting behavior. 

\subsection{Datasets}
\label{appendix:datasets}
We evaluate our design for the application of urban event reconstruction.
Three types of urban events are considered: 1) parking violations, 2) abandoned properties, and 3) unsanitary conditions. 
For abandoned properties, we collect data from the City of Newark.
The reports are collected from the SeeClickFix platform \citep{seeclick}.
This is an online platform where residents can submit complaints about various urban issues.
We collect the labels of abandoned properties from the historical records of the City of Newark, which track and monitor abandoned properties.
The data are collected from a system for the city of Newark to track and monitor the status of abandoned properties.
When an abandoned property is identified by the city agents, a record is created in the system.
Each record tracks the progress of handling an abandoned property, including inspections, rehabbing, and demolishing.
It also includes information about when the properties are detected by the city.
With detailed tracking of abandoned properties, we label the properties in the time domain to avoid mismatches between labels and multi-view data sources.
Specifically, we assigned a detection time for each property, where no events after the time are collected for all views. 
We use this information to count the number of abandoned properties during each time period. 
We collect the data from Jan. 2019 to May 2021 in Newark.

We collect data on parking violations and unsanitary issues from Manhattan in 2023.
The residents' reports are collected from the NYC311 platform. 
The ground-truth of parking violations is collected from the database of parking violation tickets provided by the department of finance \citep{data_pv}.
The ground-truth of unsanitary conditions is collected from the hearing records of the city's administrative law court \citep{data_sc}.

The cities are partitioned based on census tracts, and we collect the residents' socioeconomic status in each region from the census bureau database \citep{data_census}. 
As a result, the city of Newark is partitioned into 88 regions, and Manhattan is partitioned into 322 regions.
We count the frequency of urban events and reports every month.
The socioeconomic status factors include 5 categories, as shown in Table~\ref{tab:ses}.
There are 22 factors collected, and the details of each factor can be found in \citep{data_census}.

\begin{table}[]
    \centering
    \begin{tabular}{p{3.5cm}|p{8cm}}
    \hline
    Category & Factors \\
    \hline 
    Finance & mean income, unemployment rate, mortgage ratio, poverty rate, housing cost \\
    \hline
    Education Attainment & \% less than high school, \% high school, \% bachelor or higher\\
    \hline
    Race \& Culture & \% Hispanic, \% white, \% black, \% Asian\\
    \hline
    Access to Technology & \% has computer, \% has smartphone, \% has internet\\
    \hline
    Social Environment & population, \% multifamily, \% owner occupied, \% renter occupied, median year built, \% room occupation $\geq 0.5$, mobility rate\\
    \hline
    \end{tabular}
    \caption{An overview of the socioeconomic status factors. 
    }
    \label{tab:ses}
    \vspace{-0.2in}
\end{table}

\subsection{Implementation details}
\label{appendix:implementation}

The multi-task SEM is implemented according to Figure~\ref{fig:framework}.
Each neural network block (the white nodes) is implemented using a 4-layer MLP with a hidden dimension of 64.
For the mechanism variables, their generation is fitted with the module in Figure~\ref{fig:multi-task_P_SEM}, where each neural network block is also implemented using an MLP with a hidden dimension of 64.
As discussed in Section~\ref{sec:urban_event}, we set the number of mechanism variables as $|W|=5$ to represent the psychological pathways that affect reporting behavior.
For mechanism variables' SEMs, the first 2 layers are partitioned into the task-shared backbone and task-dependent deviation.
Training parameters are set as: $\lambda_s=10^{-3},~\lambda_d=5\cdot 10^{-3},~ \lambda_{\text{div}}=10^{-4}, ~ \lambda_{\text{wd}}=10^{-4}$ and $c=0.5$.
80\% of the data is used for training and validation, while the other 20\% is used for evaluation.
Z-score normalization is applied.
Since BSM-UR, CEVAE, and TEDVAE do not employ multi-task learning, they are trained and evaluated on each task separately.
A 10-fold experiment is conducted on a personal computer with an Intel 13700K CPU and an NVIDIA 4070 Ti SUPER GPU.
The training takes 253.2 seconds on average for each trial.

\subsection{Ablation Study}
\label{appendix:ablation}

\begin{table*}[t]
    \centering
    \small
    \setlength{\tabcolsep}{6pt}
    \begin{tabular}{c|cc|cc|cc}
    \textbf{Model} 
    & \multicolumn{2}{c|}{\textbf{Parking Violation}}
    & \multicolumn{2}{c|}{\textbf{Abandoned Property}}
    & \multicolumn{2}{c}{\textbf{Unsanitary Condition}} \\
    & \textbf{MAE} & \textbf{MSE}
    & \textbf{MAE} & \textbf{MSE}
    & \textbf{MAE} & \textbf{MSE} \\
    \hline
    Complete MTAC     & \textbf{0.2971} & \textbf{0.1616} & \textbf{0.4163} & \textbf{0.1910} & \textbf{0.3755} & \textbf{0.2501} \\
    w/o MAP         & 0.5841 & 0.6977 & 0.7841 & 0.6006 & 0.5957 & 0.7426 \\
    w/o Causality   & 0.3916 & 0.1811 & 0.4967 & 0.2842 & 0.4287 & 0.2938 \\
    
    \end{tabular}
    \caption{Ablation study. For MTAC without MAP, the cause is estimated directly from the forward causal model.}
    \label{tab:eval_ablation}
\end{table*}

To understand the contribution of each component of MTAC, we conduct an ablation study.
The results are presented in Table~\ref{tab:eval_ablation}.
We evaluate MTAC when the cause is estimated directly from the forward causal model instead of using the MAP-based inference algorithm.
The prediction error increases significantly on all three tasks.
On the abandoned property dataset, the MAE increases by 254.49\%.
This is because the causal model is a forward model, which computes the distribution of variables given its parental variables.
Direct cause estimation only uses the SES information of residents but not the reports, i.e., $p(X|Z)$ rather than $p(X|Z,Y)$.
Therefore, the direct estimation is highly inaccurate without the strongest indicator $Y$.

To estimate the gain of modeling the causal structure compared to correlational relations, we evaluate MTAC without the causal structure.
Specifically, a task-shared MLP module transfers SES variables to an embedding vector.
For each task, a task-dependent head takes the embedding vector concatenated with the report count as input and generates the estimated event count. 
This model learns the correlative relation across multiple tasks, rather than modeling the causal structure of reporting behavior.
As shown in Table~\ref{tab:eval_ablation}, the complete MTAC achieves higher reconstruction accuracy in all tasks. 
By modeling the causal structure, MTAC separates the reporting bias and true event incidence through joint inference under the assumed structure.
The gain in reconstruction accuracy highlights the significance of modeling causal structure.

\subsection{Policy Sensitivity Analysis}
\label{appendix:policy_sensitivity}



To understand MTAC's causal utility, we also evaluate MTAC's performance under distribution shift. Specifically, we evaluate the prediction accuracy in regions with various SES profiles.
We apply principal component analysis (PCA) to SES and stratify the test tracts into 4 SES quartiles based on the PC1 component.
MTAC's reconstructed true event count achieves 2.2--29 times lower MAE than leveraging raw reports as event observations across every task and quartile.
Across all three tasks and all four SES quartiles, MTAC consistently attains lower MAE than raw reports: 
for abandoned property, MTAC achieves 0.11--0.13 events/tract versus 2.2--3.7 for raw reports (ratio: 18.8--28.9$\times$), with the largest gap in the lowest-SES quartile (Q1: $0.129$ vs.\ $3.722$, ratio $28.9\times$); 
for parking violation, MTAC achieves 13.3--14.6 versus 39.7--51.0 (ratio: 2.7--3.8$\times$), with broadly stable MTAC MAE across quartiles (range: 1.3 events/tract) despite large variation in raw report counts (range: 11.3 events/tract); 
and for sanitary condition, MTAC achieves 2.4--3.4 versus 5.3--8.1 (ratio: 2.2--2.7$\times$), where the absolute gap widens monotonically from high- to low-SES tracts (Q4: $2.367$ vs.\ $5.277$; Q1: $3.358$ vs.\ $8.113$).
Raw reports show strongly SES-varying bias (under-counting by 6.8 events in low-SES vs.\ 2.6 in high-SES quartiles), while MTAC's worst-case bias across all tasks and quartiles is less than 3.4 events. 
This demonstrates causal utility beyond curve-fitting: MTAC recovers an estimator robust to confounder shift.

\subsection{Hyperparameter Sensitivity}
\label{appendix:sweep}

\begin{figure}
    \centering
    \begin{subfigure}{.45\linewidth}
        \centering
        \includegraphics[width=0.8\linewidth]{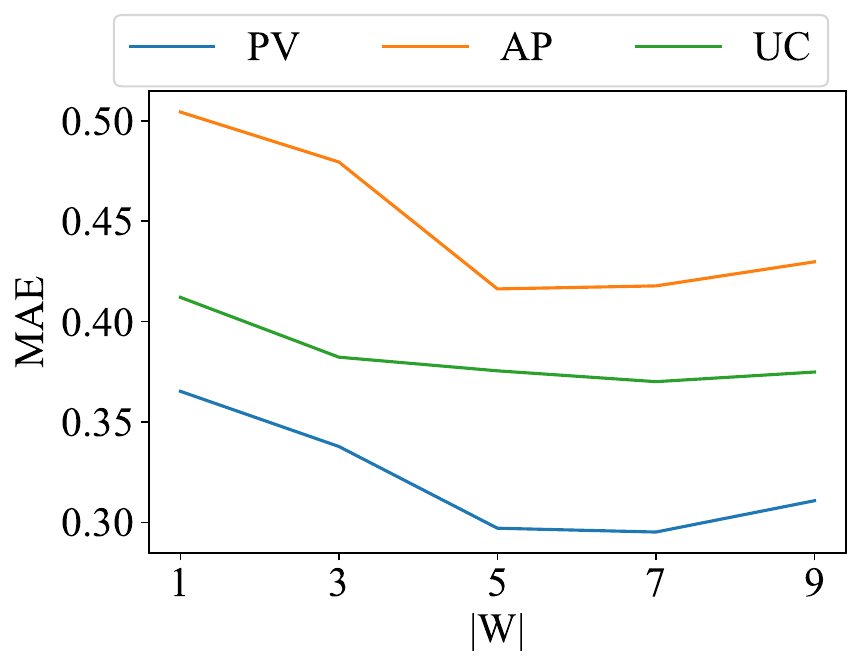}
        \caption{MAE against number of mechanism variables.}
        \label{fig:eva:nW_sweep}
    \end{subfigure}
    \hfill
    \begin{subfigure}{.45\linewidth}
        \centering
        \includegraphics[width=0.8\linewidth]{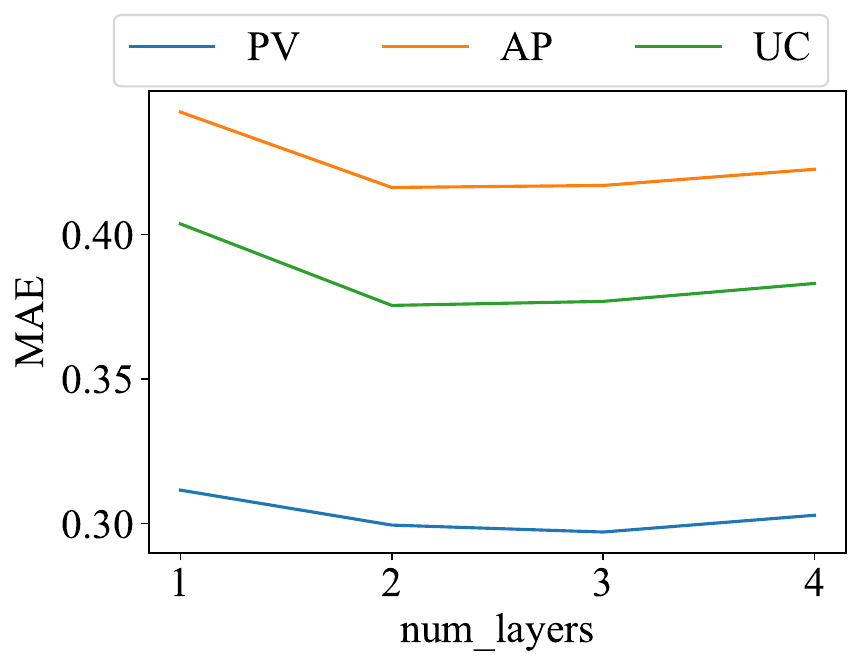}
        \caption{MAE against deviation layers.}
        \label{fig:eva:nlayer_sweep}
    \end{subfigure}
    \caption{Sensitivity of reconstruction performance on model parameters.}
    \label{fig:placeholder}
\end{figure}

As inspired by~\citep{mannarini2010invol, rotman2014partic}, there are multiple psychological paths that affect residents' reporting behavior, as discussed in Appendix~\ref{appendix:case_study}.
To understand the dimonsionality the multi-path psychological mechanism, we evaluate MTAC while sweeping the number of mechanism variables.
Figure~\ref{fig:eva:nW_sweep} illustrates the reconstruction MAE performance accordingly.
The reconstruction error is improved as the number of mechanism variables reaches 5, and stay stable afterwards. 
An overfitting phenomenon is observed with 9.
The degradation in reconstruction loss with fewer mechanism variables indicates that the reporting preference is high-dimensional.
It is a compound effect of multiple psychological paths.
As such, it is necessary for MTAC to have multiple mechanism variables to model reporting preferences.
This result shows that there exist multi-dimensional psychological path of residents' reporting behavior in~\citep{kontokosta2017equity,mannarini2010invol, Peterson2003psy,rotman2014partic}.

Figure~\ref{fig:eva:nlayer_sweep} demonstrates the reconstruction MAE with the number of layers being partitioned into task-shared backbone and task-dependent deviation.
We implement the mechanism SEM module with 4 layers.
Although only first layer serves parent selection,
extending the deviation module to more layers shows reconstruction improvement across all tasks.
This is because, besides the structural relation, the dependency effect is also task-dependent, and deviation module with only one layer is not representative enough to capture it.

\end{document}